\documentclass{article}

\usepackage{PRIMEarxiv}

\usepackage[utf8]{inputenc} % allow utf-8 input
\usepackage[T1]{fontenc}    % use 8-bit T1 fonts
\usepackage{hyperref}       % hyperlinks
\usepackage{url}            % simple URL typesetting
\usepackage{booktabs}       % professional-quality tables
\usepackage{amsfonts}       % blackboard math symbols
\usepackage{nicefrac}       % compact symbols for 1/2, etc.
\usepackage{microtype}      % microtypography
\usepackage{lipsum}
\usepackage{fancyhdr}       % header
\usepackage{graphicx}       % graphics
\graphicspath{{media/}}     % organize your images and other figures under media/ folder
\usepackage{multirow}
\usepackage{amsmath}
\usepackage{cleveref}
\usepackage{pifont}
\usepackage{xcolor}
\definecolor{myblue}{rgb}{0.21,0.49,0.74}

\hypersetup{
hidelinks,
colorlinks=true,
linkcolor=red,
citecolor=myblue
}

\title{TrajVG: 3D Trajectory-Coupled Visual Geometry Learning}

\author{
\textbf{Xingyu\,Miao$^{1}$\;\;\,
Weiguang\,Zhao$^{2}$\;\;\,
Tao\,Lu$^{3}$\;\;\,
Linning\,Xu$^{3}$\;\;\,
Mulin\,Yu$^{3}$}\\
\textbf{Yang\,Long$^{1,*}$\;\;\,
Jiangmiao\,Pang$^{3}$\;\;\,
Junting\,Dong$^{3,*,\dagger}$}\\[0.5em]
$^{1}$Durham University \quad
$^{2}$University of Liverpool \quad
$^{3}$Shanghai AI Lab \\[0.5em]
Project page: \url{https://xingy038.github.io/TrajVG/}
}

\begin{document}
\maketitle

\begingroup
\renewcommand{\thefootnote}{\fnsymbol{footnote}}
\footnotetext[2]{Project lead.}
\footnotetext[1]{Corresponding author.}
\renewcommand\thefootnote{}\footnotetext{This work was completed during an internship at Shanghai AI Lab.}
\endgroup

\begin{abstract}
Feed-forward multi-frame 3D reconstruction models often degrade on videos with object motion. Global-reference becomes ambiguous under multiple motions, while the local pointmap relies heavily on estimated relative poses and can drift, causing cross-frame misalignment and duplicated structures. We propose TrajVG, a reconstruction framework that makes cross-frame 3D correspondence an explicit prediction by estimating camera-coordinate 3D trajectories. We couple sparse trajectories, per-frame local point maps, and relative camera poses with geometric consistency objectives: (i) bidirectional trajectory–pointmap consistency with controlled gradient flow, and (ii) a pose consistency objective driven by static track anchors that suppresses gradients from dynamic regions. To scale training to in-the-wild videos where 3D trajectory labels are scarce, we reformulate the same coupling constraints into self-supervised objectives using only pseudo 2D tracks, enabling unified training with mixed supervision. Extensive experiments across 3D tracking, pose estimation, point-map reconstruction, and video depth show that TrajVG surpasses the current feedforward performance baseline.
\end{abstract}

\section{Introduction}
Geometric reconstruction from images seeks to recover camera motion and 3D structure from visual observations~\cite{hartley2003multiple,triggs1999bundle}, and it remains a core capability for systems that need spatial understanding, such as augmented reality, robotics, and navigation. For decades, the dominant approach has been geometric estimation, i.e., correspondences are extracted and used to solve for camera poses and 3D structure, with global consistency enforced by iterative refinement procedures such as bundle adjustment~\cite{lowe2004distinctive,schonberger2016structure,agarwal2011building}. Feed-forward 3D models have recently provided an alternative by learning to predict dense geometry and camera relationships directly from pixels~\cite{teed2021droid,zhou2017unsupervised}. A common representation of geometry as pixel-aligned point maps, turning reconstruction into dense regression with implicit correspondence reasoning inside the network~\cite{dust3r_cvpr24,murai2025mast3r,fang2025dens3r}. However, by prioritizing per-pixel geometry regression, these models often overlook the importance of explicitly learning cross-frame correspondences.

Recent feed-forward reconstruction methods diverge fundamentally in how they represent geometry and aggregate information across frames. Reference-frame approaches, such as VGGT~\cite{wang2025vggt}, define predictions relative to a fixed world reference (e.g., the first camera). While effective for static scenes, this rigid global anchoring often fails to represent dynamic objects or complex motions where a single reference frame is insufficient. Conversely, local-frame approaches like $\pi^3$~\cite{wang2025pi} mitigate this by predicting scale-invariant point maps in a purely per-view manner. While this design handles non-rigid motion flexibly, it introduces a critical bottleneck, i.e., the fusion of these independent local maps is mediated solely by the estimated camera pose. In reality, consecutive frames share a strong physical coupling that goes beyond simple rigid transformations. Although recent frameworks like MASt3R~\cite{murai2025mast3r}, St4RTrack~\cite{st4rtrack2025}, and VGGT capture these cross-view dependencies implicitly, they lack explicit geometric constraints to enforce this coupling. Consequently, when pose estimation is ambiguous (e.g., due to weak overlap or rotational drift), the lack of explicit tie points leads to misalignment, such as duplicated surfaces or drift, even if individual point maps are locally coherent.

Inspired by classical SfM, we elevate cross-frame 3D correspondence from an implicit byproduct of dense regression to an explicit prediction that links per-view point maps. We propose \textbf{TrajVG}, a feed-forward reconstruction framework that enforces explicit geometric coupling across frames by introducing trajectory-based 3D correspondences. Rather than depending on a fixed reference coordinate system or the camera pose head to align these maps, we introduce a 3D tracking branch that estimates camera-coordinate trajectories conditioned on both multi-frame visual features and current geometric predictions. These trajectories act as geometric tie points and drive two coupling objectives: (i) bidirectional consistency between trajectories and sampled point-map values, and (ii) pose consistency obtained by transforming trajectory points into a common reference frame.

\begin{figure*}
    \centering
    \includegraphics[width=0.9725\linewidth]{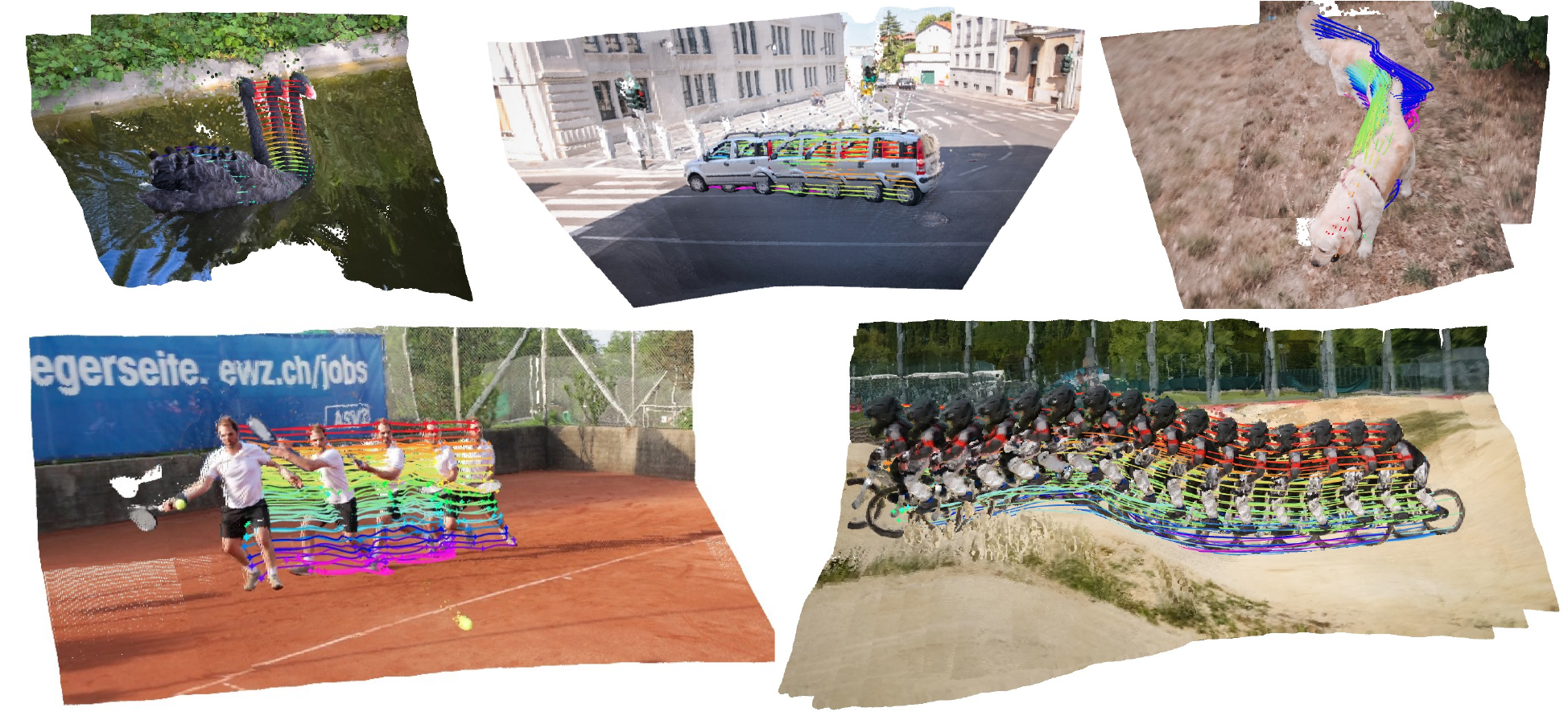}
    \caption{\textbf{Visualization results of TrajVG on in-the-wild videos.}}
    \label{fig:qualitative}
\end{figure*}

Specifically, we leverage a 3D tracking branch alongside the geometry and pose heads, where correspondences are formulated as 3D trajectories in camera space explicitly anchored to the predicted local point maps. This branch conditions not only on multi-frame visual features but also on the current geometric predictions, ensuring that matching operates within the lifted 3D space rather than in the image plane. We further transform these independent correspondence predictions into active supervision signals by constructing a set of geometric coupling constraints that tightly link sparse tracks, dense point maps, and camera poses. By employing explicit gradient routing, we prevent any single component from absorbing the entire residual error.

We enforce this coupling through distinct optimization objectives. For geometry, we establish bidirectional consistency by sampling local point maps at tracking locations and minimizing their deviation from the predicted trajectories. This objective allows high-confidence tracks to regularize the geometry while simultaneously using the geometry as evidence to refine trajectory estimates. For camera pose, we impose pose consistency by projecting these trajectory points into a common reference frame using the predicted relative poses. To ensure robust pose optimization, we introduce a static-point gating mechanism that suppresses gradients from dynamic regions. This ensures that pose updates are driven primarily by points consistent with rigid motion. To extend to real-world video where 3D trajectory annotations are scarce, we adapt these coupling constraints into a self-supervised objective dependent solely on 2D tracks. 2D tracked pixels are used to sample predicted point maps to obtain camera-frame 3D observations. This enables unified training across datasets with mixed supervision and improves generalization to in-the-wild videos. To summarize, the main contributions of this work are as follows:
\begin{itemize}
  \item We present a multi-frame reconstruction pipeline that treats 3D tracking as an explicit correspondence relationship to predict per-frame camera-coordinate trajectories to directly improve point-map fusion and relative pose estimation.

  \item We introduce two complementary consistency terms that couple trajectories, point maps, and relative poses, which force the outputs to agree across frames and stabilize optimization in dynamic scenes through static-point gating and controlled gradient flow.

  \item We introduce a semi-supervised training scheme that transfers the same coupling constraints to in-the-wild videos using pseudo 2D tracks, enabling mixed-supervision training and improving generalization.

\end{itemize}

\section{Related Work}
\subsection{Feed-Forward 3D Reconstruction}
Recent methods for 3D reconstruction from images have shifted from classical geometric optimization (e.g., bundle adjustment) to learning-based approaches, where neural networks predict dense geometry and camera relations directly from pixels. The earliest works in this area, like DUSt3R~\cite{dust3r_cvpr24}, predicted point maps by regressing from images, turning the problem of correspondence matching into a learned process. This was followed by methods like MASt3R~\cite{mast3r_eccv24} that introduced a dense feature head and matching loss to improve geometry consistency across frames.

These feed-forward models can be divided into two categories based on the coordinate system used for multi-frame fusion. One class of methods, such as VGGT~\cite{wang2025vggt}, uses a shared world reference frame, anchoring the 3D structure and camera poses to a fixed coordinate system. This approach is effective when the scene contains a stable reference but can fail when the reference is weak or when there is substantial non-rigid motion. In contrast, $\pi^3$~\cite{wang2025pi} avoids using a fixed reference view and instead employs a permutation-equivariant design, predicting scale-invariant local point maps and affine-invariant camera relations per view. This design is more robust to complex motion patterns and occlusions, as it does not force a single global explanation of the scene.

\subsection{Point Tracking}
Recently, most point tracking work targets 2D trajectories in the image plane~\cite{cotracker2, cotracker3}. PIPs~\cite{harley2022particle} address long-term tracking through occlusions. TAPIR~\cite{doersch2023tapir} tracks arbitrary queried points with per-frame initialization plus temporal refinement, and CoTracker series~\cite{cotracker2, cotracker3} improves robustness by tracking many points jointly with a transformer. These methods provide strong 2D supervision signals, but they do not directly constrain 3D alignment unless paired with depth and pose. Some feed-forward reconstruction models also output tracks as part of joint geometry prediction, for example VGGT~\cite{wang2025vggt} includes 2D point tracks in its global-frame formulation.

3D point tracking has recently become more active, supported by benchmarks such as TAPVid-3D~\cite{tapvid3d}. Early directions include optimization-heavy methods that track many points across space and time, such as OmniMotion~\cite{wang2023omnimotion}. More recent methods lift tracking into 3D by combining tracking with depth and camera. DELTA~\cite{ngo2024delta} takes RGB video together with depth maps obtained from an off-the-shelf monocular depth estimator, and then uses an attention-based tracking. In the multi-camera setting, MVTracker~\cite{rajic2025mvtracker} targets multi-view 3D point tracking using a practical number of views, and assumes known camera poses together with sensor depth or estimated multi-view depth. It fuses multi-view features into a 3D point cloud, uses KNN correlation, and iteratively refines tracks with a transformer update, improving robustness under occlusion and depth ambiguity. SpaTracker~\cite{SpatialTracker} targets spatio-temporal “track any point” in 3D. It lifts pixels to 3D with monocular depth, represents each frame in a compact 3D form, and uses iterative transformer updates to estimate 3D trajectories. TAPIP3D~\cite{tapip3d} represents videos as camera-stabilized 3D feature clouds and iteratively refines 3D motion, producing tracks in world coordinates. SpatialTrackerV2~\cite{SpatialTracker2} unifies point tracking, monocular depth, and camera pose estimation in a single feed-forward 3D tracker, producing geometry, poses, and trajectories together.

Compared with these 3D tracking lines, we treat 3D tracking as an explicit correspondence module inside a reconstruction system, and we couple tracks, dense point maps, and poses through bidirectional consistency and pose gating losses. This makes the tracking branch act as a cross-frame geometric constraint for reconstruction, rather than treating tracking as a byproduct of geometric reconstruction.

\begin{figure*}
    \centering
    \includegraphics[width=0.9275\linewidth]{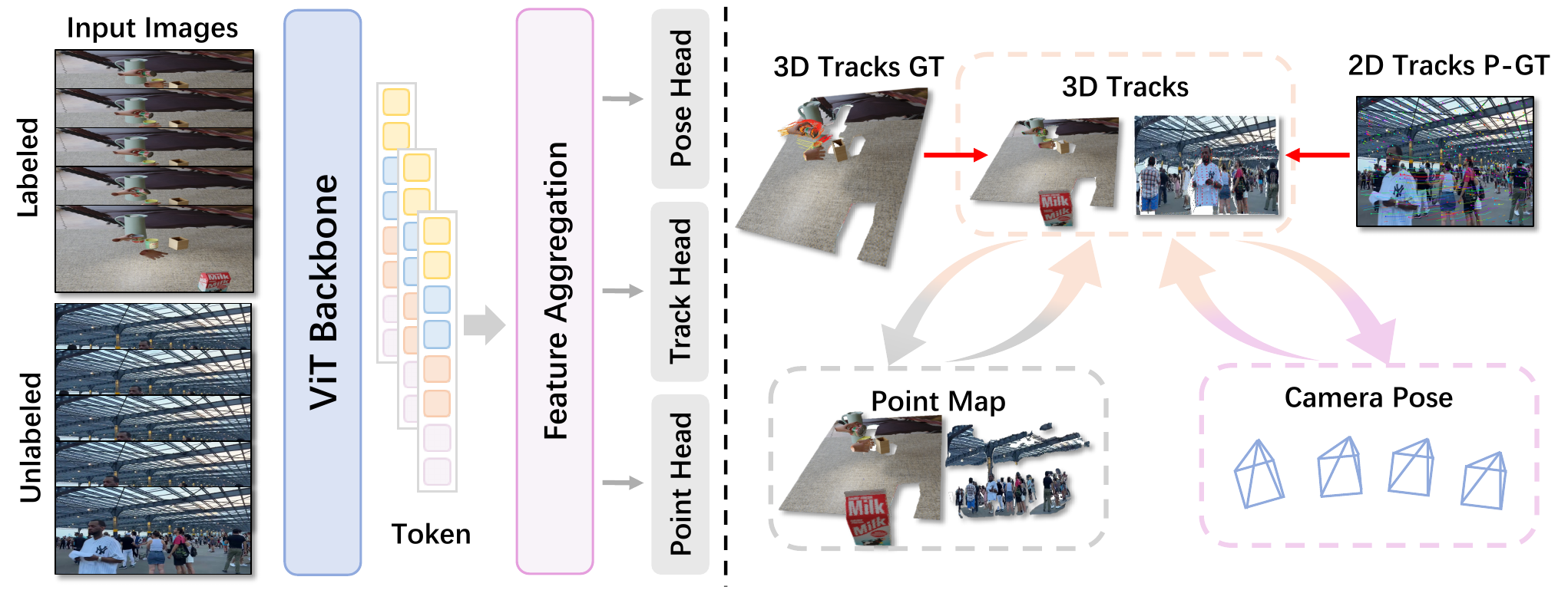}
    \vspace{-1em}
    \caption{\textbf{Overview of our model.} We jointly 3D track points with point-map and camera poses that tracking provides a direct improvement on both geometry reconstruction and camera motion.}
    \label{fig:pipeline}
\end{figure*}

\section{Method}
\label{sec:method}
In \Cref{fig:pipeline}, our model predicts per-frame local pointmaps and camera poses, and we add a 3D tracking head to learn correspondences at scale from mostly static data. We couple tracks, pointmaps, and cameras through correlation-based constraints so that tracking supervision provides gradients to both the pointmap and pose branches. When metric 3D ground truth is unavailable, we train using only 2D tracks via self-supervised variants of the same coupled objectives.

\subsection{Preliminaries}
\label{sec:prelim}
We index frames by $t\in\{0,\dots,T-1\}$ and denote the image at time $t$ as $I_t$.
Each frame has its own camera coordinate system, denoted by $c_t$.

\paragraph{\textbf{Camera poses.}}
The inverse pose maps world to camera coordinates:
$\hat{\mathbf{C}}_{w\rightarrow c_t}=\hat{\mathbf{C}}_t^{-1}$.
The relative pose from frame $t$ to an anchor frame $x$ is
\begin{equation}
\hat{\mathbf{C}}_{t\rightarrow x}=\hat{\mathbf{C}}_x^{-1}\hat{\mathbf{C}}_t .
\end{equation}

\paragraph{\textbf{Local pointmaps.}}
For each frame $t$, the model predicts a pixel-aligned local pointmap
\begin{equation}
\hat{\mathbf{P}}_t(u)\in\mathbb{R}^3, u=(x,y)\in\Omega,
\end{equation}
where $\hat{\mathbf{P}}_t(u)$ is expressed in camera coordinates $c_t$.

\subsection{3D Tracking}
\label{sec:tracking}
Recent feed-forward 3D reconstruction models are often trained on large multi-view video datasets dominated by static geometry, such as indoor RGB-D scans (e.g., ScanNet \cite{dai2017scannet}) and object-centric captures (e.g., CO3D \cite{reizenstein21co3d}). In this setting, most apparent motion is caused by ego-motion rather than scene deformation, so 2D pixel-space tracking supervision is well-posed. Even when the underlying 3D point is static, its image-plane location changes with viewpoint, so the model must use cross-frame appearance and occlusion cues to maintain consistent tracks (e.g., VGGT \cite{wang2025vggt}).

Our goal is to learn 3D correspondences that mutually improve both pointmaps and camera poses. However, if trajectories are defined in a fixed world (or canonical) coordinate system, static points have a trivial solution, that is, for a static world point $\mathbf{X}_i$, the correct trajectory is time-invariant, i.e., $\mathbf{X}_{t,i}=\mathbf{X}_i$. In this scenario, the 3D tracking branch can minimize its loss by simply producing near-zero updates to maintain a constant state, effectively ignoring the specific local correspondence in frame $t$. As the state becomes constant, the local geometric neighborhoods and correlation features used by the update rule also become similar across time, which further encourages a near-zero update rule and makes the tracking gradients insensitive to correspondence errors. This weakens the signals needed for feed-forward reconstruction, such as correcting pointmap overlap (duplicate surfaces), improving cross-frame alignment, and constraining relative camera motion.

We avoid this by predicting tracks in the camera coordinate system of each frame. For a static world point $\mathbf{X}_i$, its camera coordinate position:
\begin{equation}
\mathbf{p}^{(c_t)}_{t,i}=\hat{\mathbf{C}}_{w\rightarrow c_t}\,\mathrm{Hom}(\mathbf{X}_i).
\end{equation}
where $\mathrm{Hom}(\mathbf{x})=[\mathbf{x}^\top,1]^\top$ is homogeneous coordinates. The tracking branch must explain viewpoint-induced geometric changes, and must re-associate the query point with the correct local neighborhood on each frame's predicted pointmap.

Given query pixels $\{\mathbf{q}_i\}_{i=1}^{N_q}$ in the first frame, we initialize the 3D query point by sampling the first-frame pointmap:
\begin{equation}
\hat{\mathbf{p}}_{0,i}=\mathrm{Sample}\!\left(\hat{\mathbf{P}}_0,\mathbf{q}_i\right).
\label{sample}
\end{equation}
where $\mathrm{Sample}(\hat{\mathbf{P}}_t,u)$ denotes bilinear sampling on the image grid. The 3D tracking branch then predicts a camera coordinate trajectory $\hat{\mathbf{p}}_{t,i}\in\mathbb{R}^3$ and visibility $\hat{v}_{t,i}\in[0,1]$ for each frame.

\subsection{Joint pointmaps, tracks, and cameras optimization.}
\label{sec:coupled_loss}
If the 3D tracking head is trained only to match trajectory labels, it may remain an auxiliary prediction that may not fix pointmap overlap or reduce pose drift. We therefore introduce coupling terms that connect (i) tracked 3D points, (ii) pointmap values at tracked pixels, and (iii) predicted relative camera motion.

Let $\mathbf{q}_{t,i}$ denote the 2D location of query $i$ in frame $t$, and define the pointmap sample (in $c_t$) as:
\begin{equation}
\tilde{\mathbf{p}}_{t,i} = \mathrm{Sample}\!\left(\hat{\mathbf{P}}_t,\mathbf{q}_{t,i}\right)\in\mathbb{R}^3.
\end{equation}
We use a visibility weight $w_{t,i}$ and a Huber penalty $\rho(\cdot)$.

\paragraph{\textbf{Tracking-pointmap bidirectional consistency.}}
We enforce agreement between the tracked 3D point and the pointmap value at the tracked pixel, while controlling which branch receives gradients:
\begin{equation}
\mathcal{L}_{\mathrm{cons}}
=
\sum_{t,i} w_{t,i}\Big(
\rho\!\left(\mathrm{sg}\!\left[\hat{\mathbf{p}}_{t,i}\right]-\tilde{\mathbf{p}}_{t,i}\right)
+
\rho\!\left(\hat{\mathbf{p}}_{t,i}-\mathrm{sg}\!\left[\tilde{\mathbf{p}}_{t,i}\right]\right)
\Big).
\label{eq:cons_bidir}
\end{equation}

where $\mathrm{sg}[\cdot]$ is the stop-gradient operator.The first term updates the pointmap branch (tracks detached), while the second updates the tracking branch (pointmap samples detached). This avoids unstable gradient interactions where the two branches chase each other, and it also avoids a degenerate compromise where both outputs drift toward an averaged value.

\begin{figure*}
    \centering
    \includegraphics[width=0.9725\linewidth]{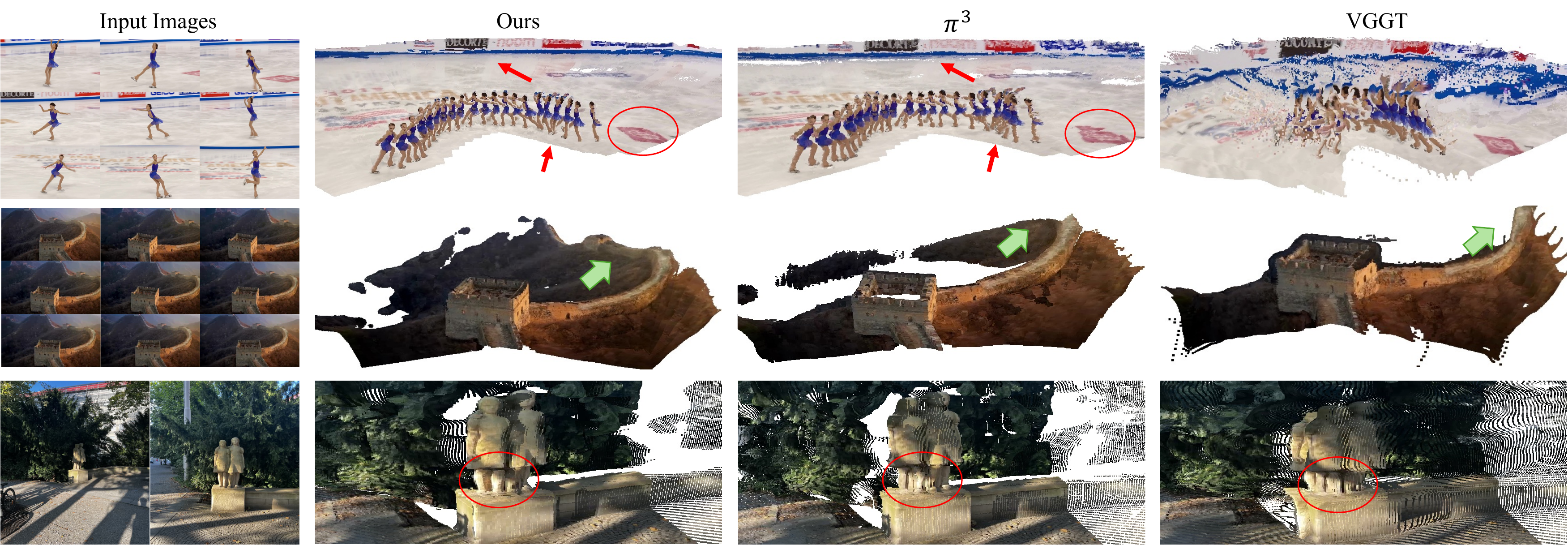}
    \caption{\textbf{Qualitative comparison of multi-view 3D reconstruction.} Our method achieves better reconstruction results in field scenarios. In contrast, baseline methods suffer from issues such as overlap and loss of detail.}
    \label{fig:qualitative}
\end{figure*}

\paragraph{\textbf{Camera consistency from track anchors.}}
To accurately recover ego-motion, we must isolate rigid background geometry from dynamic objects. We therefore define a binary static mask $m_{t,i}$ using ground-truth 3D trajectories. Let $\mathbf{X}^*_{t,i}$ and $\mathbf{X}^*_{x,i}$ denote the ground-truth world coordinates of query $i$ at frame $t$ and anchor frame $x$, respectively. A track is considered static (i.e., $m_{t,i}=1$) only if its physical displacement satisfies $\|\mathbf{X}^*_{t,i} - \mathbf{X}^*_{x,i}\|_2 < \tau_{\mathrm{static}}$. In implementation, we compute this mask in the anchor camera coordinates using a robust temporal reference (e.g., the median over time) to reduce sensitivity to outliers.

Under the predicted relative motion, these static points should map to consistent locations in the anchor camera coordinates $c_x$. We express a per-frame 3D point in $c_x$ via:
\begin{equation}
\tilde{\mathbf{p}}^{(c_x)}_{t,i} = \hat{\mathbf{C}}_{t\rightarrow x}\,\mathrm{Hom}\!\left(\mathrm{sg}\!\left[\tilde{\mathbf{p}}_{t,i}\right]\right).
\end{equation}

We then enforce (i) anchor consistency between reprojected static samples and the anchor-frame samples, and (ii) supervision against the ground-truth anchor targets $\bar{\mathbf{p}}^{(c_x)}_{t,i}$:
\begin{equation}
\mathcal{L}_{\mathrm{cam}}
=
\sum_{t,i} w_{t,i}\Big(
m_{t,i}\,
\rho\!\left(
\hat{\mathbf{C}}_{t\rightarrow x}\,
\mathrm{Hom}\!\left(\mathrm{sg}\!\left[\hat{\mathbf{p}}_{t,i}\right]\right)
-\bar{\mathbf{p}}^{(c_x)}_{t,i}\right)
\nonumber
+
\rho\!\left(
\mathrm{sg}\!\left[\hat{\mathbf{C}}_{t\rightarrow x}\right]\,
\mathrm{Hom}(\hat{\mathbf{p}}_{t,i})
-\bar{\mathbf{p}}^{(c_x)}_{t,i}\right)
\Big).
\label{eq:cam_anchor}
\end{equation}

where $\mathrm{sg}[\cdot]$ controls the update direction: the first term updates only the pose $\hat{\mathbf{C}}_{t\rightarrow x}$ (using static tracks), while the second term updates only the tracked 3D points.

\begin{table}
    \centering
    \caption{\textbf{Comparison of 3D point tracking methods on the large-scale real-world TAPVid-3D~\cite{tapvid3d} benchmark.} C, U, and M are abbreviations for COLMAP \cite{schonberger2016structure}, Unidepth \cite{piccinelli2024unidepth}, and MegaSAM \cite{li2025megasam}. $^{\dagger}$ denotes using depth to lift 2D tracks to 3D tracks.}
    \resizebox{\linewidth}{!}{
    \begin{tabular}{l|ccc|ccc|ccc}
        \toprule
        \multirow{2}{*}{\textbf{Methods}}& \multicolumn{3}{c|}{\textbf{ADT}} & \multicolumn{3}{c|}{\textbf{DriveTrack}} & \multicolumn{3}{c}{\textbf{PStudio}} \\
        & AJ$_{\text{3D}}\uparrow$ & APD$_{\text{3D}}\uparrow$ & OA$\uparrow$ & AJ$_{\text{3D}}\uparrow$ & APD$_{\text{3D}}\uparrow$ & OA$\uparrow$  & AJ$_{\text{3D}}\uparrow$ & APD$_{\text{3D}}\uparrow$ & OA$\uparrow$  \\ \midrule
        TAPIR$^{\dagger}$~\cite{doersch2023tapir} + C & 7.1 & 11.9 & 72.6 & 8.9 & 14.7 & 80.4 & 6.1 & 10.7 & 75.2 \\
        CoTracker3$^{\dagger}$~\cite{cotracker3} + C & 8.0 & 12.3 & 78.6 & 11.7 & 19.1 & 81.7 & 8.1 & 13.5 & 77.2 \\
        \midrule
        CoTracker3$^{\dagger}$~\cite{cotracker3} + U & 15.8 & 24.4 & 88.9 & 13.5 & 19.9 & 87.1 & 9.2 & 13.8 & 84.2 \\
        TAPIR$^{\dagger}$~\cite{doersch2023tapir} + U & 15.7 & 24.2 & 87.8 & 12.4 & 19.1 & 84.8 & 7.3 & 13.5 & 84.3 \\
        SceneTracker + U & - & 23.1 & - & - & 6.8 & - & - & 12.7 & - \\
        SpatialTracker~\cite{SpatialTracker} + U & 13.6 & 20.9 & 90.5 & 8.3 & 14.5 & 82.8 & 8.0 & 15.0 & 75.8 \\
        SpatialTracker2~\cite{SpatialTracker2} + U & 18.6 & 26.3 & 90.8 & \underline{16.4} & \underline{24.3} & \underline{90.2} & 18.1 & 27.6 & \underline{86.7} \\
        DELTA~\cite{ngo2024delta} + U & 16.6 & 24.4 & 86.8 & 14.6 & 22.5 & 85.8 & 8.2 & 15.0 & 76.4 \\
        DELTA2~\cite{delta2} + U & 17.0 & 24.7 & 87.2 & 15.6 & 23.8 & 84.6 & 7.7 & 14.4 & 74.1 \\
        \midrule
        CoTracker3$^{\dagger}$~\cite{cotracker3} + M & 20.4 & 30.1 &  89.8 & 14.1 & 20.3 & 88.5 & 17.4 & 27.2 & 85.0 \\
        SpatialTracker~\cite{SpatialTracker} + M & 15.9 & 23.8 & 90.1 & 7.7 & 13.5 & 85.2 & 15.3 & 25.2 & 78.1 \\
        SpatialTracker2~\cite{SpatialTracker2} + M & \underline{22.3} & \underline{32.2} & \underline{93.7} & 15.8 & 23.0 & 90.0 & \underline{18.2} & \underline{28.6} & \textbf{87.3} \\
        DELTA~\cite{ngo2024delta} + M & 21.0 & 29.3 & 89.7 & 14.6 & 22.2 & 88.1 & 17.7 & 27.3 & 81.4 \\
        TAPIP3D~\cite{tapip3d} + M & 21.6 & 31.0 & 90.4 & 14.6 & 21.3 & 82.2 & 18.1 & 27.7 & 85.5 \\
        \textbf{Our} + M & \textbf{23.1} & \textbf{32.5}  &  \textbf{93.8}  &  \textbf{17.8} & \textbf{25.2}  & \textbf{92.2} & \textbf{18.2}  & \textbf{28.8}  &  86.3 \\
        \bottomrule
    \end{tabular}}
    \label{tab:tapvid3d}
\end{table}

\begin{table}
    \centering
    \caption{\textbf{Camera Pose Estimation on RealEstate10K~\cite{zhou2018stereo} and Co3Dv2~\cite{reizenstein21co3d}.} Metrics measure the ratio of angular accuracy of rotation/translation under an error of 30 degrees, the higher the better.}
    \begin{tabular}{lcccccc}
        \toprule
        \multicolumn{1}{l}{\multirow{3}{*}{\textbf{Method}}} &
        \multicolumn{3}{c}{\textbf{RealEstate10K}} &
        \multicolumn{3}{c}{\textbf{Co3Dv2}} \\
        \cmidrule(r){2-4} \cmidrule(r){5-7}
        \multicolumn{1}{c}{} &
        RRA@30 $\uparrow$ & RTA@30 $\uparrow$ & AUC@30 $\uparrow$ &
        RRA@30 $\uparrow$ & RTA@30 $\uparrow$ & AUC@30 $\uparrow$ \\
        \midrule
        Fast3R~\cite{Yang_2025_Fast3R} 
        & 99.05 & 81.86 & 61.68 
        & 97.49 & 91.11 & 73.43\\

        CUT3R~\cite{wang2025continuous} 
        & 99.82 & 95.10 & 81.47 
        & 96.19 & 92.69 & 75.82 \\

        FLARE~\cite{zhang2025FLARE} 
        & 99.69 & 95.23 & 80.01 
        & 96.38 & 93.76 & 73.99 \\

        VGGT~\cite{wang2025vggt} 
        & 99.97 & 93.13 & 77.62 
        & 98.96 & \underline{97.13} & \textbf{88.59} \\

        $\pi^3$~\cite{wang2025pi} 
        & \underline{99.99} & \textbf{95.62} & \underline{85.90} 
        & \underline{99.05} & \textbf{97.33} & \underline{88.41} \\

        \textbf{Our} 
        & \textbf{99.99} & \underline{95.58} & \textbf{86.60} 
        & \textbf{99.14} & 96.80 & 86.00 \\
        \bottomrule
    \end{tabular}
    \label{tab:relpose-vggt}
\end{table}

\subsection{Self-Supervision from In-the-Wild Videos}
\label{sec:selfsup_2d}
A key limitation of feed-forward 3D reconstruction is the scarcity of data with reliable 3D labels. While synthetic or scanned datasets provide metric ground truth, they lack the diverse appearance, motion patterns, and occlusions found in real-world footage. To leverage the abundance of Internet videos, we extend our coupled objectives to a self-supervised setting using pseudo-ground-truth 2D trajectories.

\paragraph{\textbf{Pseudo 2D tracks.}}
For an unlabeled video clip $\{I_t\}_{t=0}^{T-1}$, we generate dense 2D trajectories and visibility flags using an off-the-shelf tracker (e.g., CoTracker3):
\begin{equation}
\mathbf{q}_{t,i}\in\mathbb{R}^2, \bar v_{t,i}\in[0,1].
\end{equation}
These 2D tracks serve as the correspondence signal for our 3D branches.

\paragraph{\textbf{Self-supervised term.}}
Instead of defining new loss functions, we adapt the consistency objectives from \cref{sec:coupled_loss} to use the pseudo-tracks. We sample the predicted pointmaps $\hat{\mathbf{P}}_t$ at the tracked locations $\mathbf{q}_{t,i}$ to obtain 3D samples $\tilde{\mathbf{p}}_{t,i}$. We then apply the bidirectional consistency terms ($\mathcal{L}_{\mathrm{cons}}$) and the anchor consistency term ($\mathcal{L}_{\mathrm{cam}}$) exactly as defined in (\cref{eq:cons_bidir} and \cref{eq:cam_anchor}), substituting the tracker-derived locations and visibility weights. Note that the second term in \cref{eq:cam_anchor} is not used because there is no ground truth for the camera pose.

This formulation transfers the robust 2D correspondence signal from the video tracker to our 3D pointmap and camera branches without requiring explicit depth or pose labels. The gradient routing remains identical: consistency losses refine the alignment between the tracking head and pointmaps, while anchor consistency drives the learning of relative camera poses.

\paragraph{\textbf{Semi-supervised training strategy.}}
While self-supervision scales correspondence learning, it cannot fully constrain metric structure. Relying solely on 2D consistency can lead to geometric drift (e.g., scale ambiguity) or degenerate solutions where the model satisfies constraints via ``shortcuts'' rather than valid 3D geometry. To address this, we employ a semi-supervised approach, training on a mixture of labeled data and in-the-wild videos. The labeled data anchors the metric scale and stabilizes learning, while the large-scale unlabeled data enhances the model's robustness to diverse appearances and complex motions.

\section{Experiments}
\subsection{Training Datasets.}
We adopt a three-stage training strategy. Our model is not trained from scratch. We initialize it with the pretrained backbone weights from $\pi^3$ and keep the backbone frozen during the first stage. In this stage, we use CO3DV2~\cite{reizenstein21co3d}, Mapfree~\cite{arnold2022mapfree}, TartanAir~\cite{tartanair2020iros}, ASE~\cite{avetisyan2024scenescript}, VKITTI~\cite{cabon2020vkitti2}, MVSynth~\cite{DeepMVS}, ARKitScenes~\cite{dehghan2021arkitscenes}, BlendedMVS~\cite{yao2020blendedmvs}, DL3DV~\cite{ling2024dl3dv}, ScanNet++~\cite{yeshwanthliu2023scannetpp}, ScanNet~\cite{dai2017scannet}, MegaDepth~\cite{MegaDepthLi18}, Waymo~\cite{Waymo}, WildRGBD~\cite{xia2024rgbd}, GTASfm~\cite{GTASfm}, HyperSim~\cite{Hypersim}, OmniWorld~\cite{zhou2025omniworld}, UnReal4K~\cite{aleotti2021neural}, MatrixCity~\cite{li2023matrixcity}, Spring~\cite{Mehl2023_Spring}, Kubric~\cite{greff2021kubric}, PointOdyssey~\cite{zheng2023point}, and DynamicReplica~\cite{DynamicReplica} to train the 3D tracking branch in the camera coordinate system. In the second stage, we mix the above datasets with Sekai. Using this combined data, we finetune the backbone and jointly train the point map branch and the camera branch. The third stage follows a setting similar to $\pi^3$~\cite{wang2025pi}. We freeze the backbone, the point map branch, and the camera branch. We then use the same datasets together with an internal dynamic 3D tracking dataset to train the confidence module and the 3D tracking branch.

\begin{table}[t]
    \centering
    \caption{
        \textbf{Camera Pose Estimation on Sintel~\cite{Butler}, TUM-dynamics~\cite{6385773} and ScanNet~\cite{dai2017scannet}.}  Metrics measure the distance error of rotation/translation, the lower the better.}
    \resizebox{1.0\linewidth}{!}{
    \begin{tabular}{lccccccccc}
        \toprule
        {\multirow{3}{*}{\textbf{Method}}} &
        \multicolumn{3}{c}{\textbf{Sintel}} &
        \multicolumn{3}{c}{\textbf{TUM-dynamics}} &
        \multicolumn{3}{c}{\textbf{ScanNet}} \\
        \cmidrule(r){2-4} \cmidrule(r){5-7} \cmidrule(r){8-10}
        &
        ATE$\downarrow$ & RPE trans$\downarrow$ & RPE rot$\downarrow$ &
        ATE$\downarrow$ & RPE trans$\downarrow$ & RPE rot$\downarrow$ &
        ATE$\downarrow$ & RPE trans$\downarrow$ & RPE rot$\downarrow$ \\
        \midrule
        Fast3R~\cite{Yang_2025_Fast3R} 
        & 0.371 & 0.298 & 13.75 
        & 0.090 & 0.101 & 1.425 
        & 0.155 & 0.123 & 3.491 \\

        CUT3R~\cite{wang2025continuous} 
        & 0.217 & 0.070 & 0.636 
        & 0.047 & 0.015 & 0.451 
        & 0.094 & 0.022 & 0.629 \\

        Aether~\cite{aether} 
        & 0.189 & 0.054 & 0.694 
        & 0.092 & 0.012 & 1.106 
        & 0.176 & 0.028 & 1.204 \\

        FLARE~\cite{zhang2025FLARE} 
        & 0.207 & 0.090 & 3.015 
        & 0.026 & 0.013 & 0.475 
        & 0.064 & 0.023 & 0.971 \\

        VGGT~\cite{wang2025vggt} 
        & 0.167 & 0.062 & 0.491 
        & \underline{0.012} & 0.010 & \underline{0.311} 
        & 0.035 & 0.015 & 0.382 \\

        $\pi^3$~\cite{wang2025pi}  
        & \textbf{0.074} & \underline{0.040} & \underline{0.282} 
        & 0.014 & \underline{0.009} & 0.312 
        & \underline{0.031} & \underline{0.013} & \textbf{0.347} \\

        \textbf{Our}  
        & \underline{0.108} & \textbf{0.038} & \textbf{0.274} 
        & \textbf{0.011} & \textbf{0.008} & \textbf{0.307} 
        & \textbf{0.030} & \textbf{0.013} & \underline{0.351} \\
        \bottomrule
    \end{tabular}
    }
    \label{tab:relpose-cut3r}
\end{table}

\begin{table}[t]
    \centering
    \caption{
        \textbf{Point Map Estimation on DTU~\cite{6909453} and ETH3D~\cite{schoeps2017cvpr}.} 
        Keyframes are selected every 5 images.
    }
    \resizebox{1.0\linewidth}{!}{
    \begin{tabular}{lccccccccccccc}
        \toprule[0.17em]
        {\multirow{4}{*}{\textbf{Method}}} &
        \multicolumn{6}{c}{\textbf{DTU}} &
        \multicolumn{6}{c}{\textbf{ETH3D}} \\
        \cmidrule(r){2-7} \cmidrule(r){8-13}
        &
        \multicolumn{2}{c}{Acc. $\downarrow$}  &
        \multicolumn{2}{c}{Comp. $\downarrow$} &
        \multicolumn{2}{c}{N.C. $\uparrow$}     & 
        \multicolumn{2}{c}{Acc. $\downarrow$}  &
        \multicolumn{2}{c}{Comp. $\downarrow$} &
        \multicolumn{2}{c}{N.C. $\uparrow$}     & \\
        \cmidrule(r){2-3} \cmidrule(r){4-5} \cmidrule(r){6-7}
        \cmidrule(r){8-9} \cmidrule(r){10-11} \cmidrule(r){12-13}
        &
        Mean & Med. &
        Mean & Med. &
        Mean & Med. &
        Mean & Med. &
        Mean & Med. &
        Mean & Med. & \\
        \midrule
        Fast3R~\cite{Yang_2025_Fast3R} 
        & 3.340 & 1.919 & 2.929 & 1.125 & 0.671 & 0.755
        & 0.832 & 0.691 & 0.978 & 0.683 & 0.667 & 0.766 \\

        CUT3R~\cite{wang2025continuous} 
        & 4.742 & 2.600 & 3.400 & 1.316 & \underline{0.679} & 0.764
        & 0.617 & 0.525 & 0.747 & 0.579 & 0.754 & 0.848 \\

        FLARE~\cite{zhang2025FLARE} 
        & 2.541 & 1.468 & 3.174 & 1.420 & \textbf{0.684} & \textbf{0.774}
        & 0.464 & 0.338 & 0.664 & 0.395 & 0.744 & 0.864 \\

        VGGT~\cite{wang2025vggt} 
        & \underline{1.338} & 0.779 & \underline{1.896} & 0.992 & 0.676 & 0.766
        & 0.280 & 0.185 & 0.305 & 0.182 & 0.853 & 0.950 \\

        $\pi^3$~\cite{wang2025pi}  
        & \textbf{1.198} & \textbf{0.646} & \textbf{1.849} & \textbf{0.607} & 0.678 & 0.768
        & \underline{0.194} & \underline{0.131} & \underline{0.210} & \underline{0.128} & \underline{0.883} & \underline{0.969} \\

        \textbf{Our}  
        & 1.435 & \underline{0.762} & 2.119 & \underline{0.621} & 0.678 & \underline{0.769}
        & \textbf{0.173} & \textbf{0.117} & \textbf{0.200} & \textbf{0.124} & \textbf{0.887} & \textbf{0.970} \\
        \bottomrule
    \end{tabular}
    }
    \label{tab:mv_recon_vggt}
\end{table}

\begin{table}[t]
    \centering
    \caption{
        \textbf{Point Map Estimation on 7-Scenes~\cite{shotton2013scene} and NRGBD~\cite{azinovic2022neural}.} 
        Keyframes are selected every 200 images (for 7-Scenes) and 500 images (for NRGBD) for \textit{sparse} view, and every 40 images (for 7-Scenes) and 100 images (for NRGBD) for \textit{dense} view.
    }
    \resizebox{1.0\linewidth}{!}{
    \begin{tabular}{lcccccccccccccc}
        \toprule[0.17em]
        {\multirow{4}{*}{\textbf{Method}}} &
        {\multirow{4}{*}{\textbf{View}}} &
        \multicolumn{6}{c}{\textbf{7-Scenes}} &
        \multicolumn{6}{c}{\textbf{NRGBD}} \\
        \cmidrule(r){3-8} \cmidrule(r){9-14}
        & &
        \multicolumn{2}{c}{Acc. $\downarrow$}  &
        \multicolumn{2}{c}{Comp. $\downarrow$} &
        \multicolumn{2}{c}{NC. $\uparrow$}     & 
        \multicolumn{2}{c}{Acc. $\downarrow$}  &
        \multicolumn{2}{c}{Comp. $\downarrow$} &
        \multicolumn{2}{c}{NC. $\uparrow$}     & \\
        \cmidrule(r){3-4} \cmidrule(r){5-6} \cmidrule(r){7-8}
        \cmidrule(r){9-10} \cmidrule(r){11-12} \cmidrule(r){13-14}
        & &
        Mean & Med. &
        Mean & Med. &
        Mean & Med. &
        Mean & Med. &
        Mean & Med. &
        Mean & Med. & \\
        \midrule

        % ================= Sparse =================
        Fast3R~\cite{Yang_2025_Fast3R} & \multirow{6}{*}{\textit{sparse}} &
        0.095 & 0.065 & 0.144 & 0.089 & 0.673 & 0.759 &
        0.135 & 0.091 & 0.163 & 0.104 & 0.759 & 0.877 \\

        CUT3R~\cite{wang2025continuous} & &
        0.093 & 0.049 & 0.102 & 0.051 & 0.704 & 0.805 &
        0.104 & 0.041 & 0.079 & 0.031 & 0.822 & 0.968 \\

        FLARE~\cite{zhang2025FLARE} & &
        0.085 & 0.057 & 0.145 & 0.107 & 0.696 & 0.780 &
        0.053 & 0.024 & 0.051 & 0.025 & 0.877 & 0.988 \\

        VGGT~\cite{wang2025vggt} & &
        \underline{0.044} & \underline{0.025} & \underline{0.056} & \underline{0.033} & 0.733 & \underline{0.845} & 0.051 & 0.029 & 0.066 & 0.038 & 0.890 & 0.981 \\

        $\pi^3$~\cite{wang2025pi} & &
        0.047 & 0.029 & 0.075 & 0.049 & \underline{0.742} & 0.841 &
        \textbf{0.026} & \textbf{0.015} & \textbf{0.028} & \textbf{0.014} & \textbf{0.916} & \underline{0.992} \\

        \textbf{Our} & &
        \textbf{0.037} & \textbf{0.022} & \textbf{0.052} & \textbf{0.033} & \textbf{0.763} & \textbf{0.874} &
        \underline{0.029} & \underline{0.018} & \underline{0.032} & \underline{0.016} & \underline{0.910} & \textbf{0.992} \\

        \midrule

        % ================= Dense =================
        Fast3R~\cite{Yang_2025_Fast3R} & \multirow{6}{*}{\textit{dense}} &
        0.040 & 0.017 & 0.056 & 0.018 & 0.644 & 0.725 &
        0.072 & 0.030 & 0.050 & 0.016 & 0.790 & 0.934 \\

        CUT3R~\cite{wang2025continuous} & &
        0.023 & 0.010 & 0.027 & \textbf{0.008} & 0.669 & 0.764 &
        0.086 & 0.037 & 0.048 & 0.017 & 0.800 & 0.953 \\

        FLARE~\cite{zhang2025FLARE} & &
        0.019 & 0.007 & 0.026 & 0.013 & 0.684 & 0.785 &
        0.023 & 0.011 & 0.018 & 0.008 & 0.882 & 0.986 \\

        VGGT~\cite{wang2025vggt} & &
        0.022 & 0.007 & 0.026 & 0.012 & 0.666 & 0.760 &
        0.017 & 0.010 & 0.015 & 0.005 & \underline{0.893} & \textbf{0.988} \\

        $\pi^3$~\cite{wang2025pi} & &
        \underline{0.016} & \underline{0.007} & \underline{0.022} & 0.011 & \underline{0.689} & \underline{0.792} &
        \underline{0.015} & \underline{0.008} & \textbf{0.013} & \underline{0.005} & \textbf{0.898} & \underline{0.987} \\

        \textbf{Our} & &
        \textbf{0.015} & \textbf{0.007} & \textbf{0.022} & \underline{0.010} & \textbf{0.691} & \textbf{0.797} &
        \textbf{0.013} & \textbf{0.007} & \underline{0.014} & \textbf{0.005} & 0.858 & 0.981 \\

        \bottomrule
    \end{tabular}
    }
    \label{tab:mv_recon_cut3r}
\end{table}

\begin{table}
    \centering
    \caption{
        \textbf{Video Depth Estimation on Sintel~\cite{Butler}, Bonn~\cite{palazzolo2019refusion} and KITTI~\cite{geiger2013vision}.}
    }
    \begin{tabular}{lccccccc}
        \toprule
        \multirow{3}{*}{\textbf{Method}} &
        \multirow{3}{*}{\textbf{Align}} &
        \multicolumn{2}{c}{\textbf{Sintel}} &
        \multicolumn{2}{c}{\textbf{Bonn}} &
        \multicolumn{2}{c}{\textbf{KITTI}} \\
        \cmidrule(r){3-4} \cmidrule(r){5-6} \cmidrule(r){7-8}
        & &
        Abs Rel $\downarrow$ & $\delta<1.25$ $\uparrow$ &
        Abs Rel $\downarrow$ & $\delta<1.25$ $\uparrow$ &
        Abs Rel $\downarrow$ & $\delta<1.25$ $\uparrow$ \\
        \midrule
        DUSt3R~\cite{dust3r_cvpr24} & \multirow{10}{*}{scale} & 0.662 & 0.434 & 0.151 & 0.839 & 0.143 & 0.814 \\
        MASt3R~\cite{mast3r_eccv24} &  & 0.558 & 0.487 & 0.188 & 0.765 & 0.115 & 0.848 \\
        MonST3R~\cite{zhang2024monst3r} &  & 0.399 & 0.519 & 0.072 & 0.957 & 0.107 & 0.884 \\
        Fast3R~\cite{Yang_2025_Fast3R} &  & 0.638 & 0.422 & 0.194 & 0.772 & 0.138 & 0.834 \\
        MVDUSt3R~\cite{tang2024mv} &  & 0.805 & 0.283 & 0.426 & 0.357 & 0.456 & 0.342 \\
        CUT3R~\cite{wang2025continuous} &  & 0.417 & 0.507 & 0.078 & 0.937 & 0.122 & 0.876 \\
        Aether~\cite{aether} &  & 0.324 & 0.502 & 0.273 & 0.594 & 0.056 & 0.978 \\
        FLARE~\cite{zhang2025FLARE} &  & 0.729 & 0.336 & 0.152 & 0.790 & 0.356 & 0.570 \\
        VGGT~\cite{wang2025vggt} &  & 0.299 & 0.638 & 0.057 & 0.966 & 0.062 & 0.969 \\
        $\pi^3$~\cite{wang2025pi} &  & \underline{0.233} & \underline{0.664} & \underline{0.049} & \underline{0.975} & \textbf{0.038} & \textbf{0.986} \\
        \textbf{Our} &  & \textbf{0.220} & \textbf{0.717} & \textbf{0.036} & \textbf{0.979} & \underline{0.040} & \underline{0.984} \\
        \midrule
        DUSt3R~\cite{dust3r_cvpr24} &
        \multirow{10}{*}{\shortstack[c]{scale \\ \& \\ shift}} &
        0.570 & 0.493 & 0.152 & 0.835 & 0.135 & 0.818 \\
        MASt3R~\cite{mast3r_eccv24} &  & 0.480 & 0.517 & 0.189 & 0.771 & 0.115 & 0.849 \\
        MonST3R~\cite{zhang2024monst3r} &  & 0.402 & 0.526 & 0.070 & 0.958 & 0.098 & 0.883 \\
        Fast3R~\cite{Yang_2025_Fast3R} &  & 0.518 & 0.486 & 0.196 & 0.768 & 0.139 & 0.808 \\
        MVDUSt3R~\cite{tang2024mv} &  & 0.619 & 0.332 & 0.482 & 0.357 & 0.401 & 0.355 \\
        CUT3R~\cite{wang2025continuous} &  & 0.534 & 0.558 & 0.075 & 0.943 & 0.111 & 0.883 \\
        Aether~\cite{aether} &  & 0.314 & 0.604 & 0.308 & 0.602 & 0.054 & 0.977 \\
        FLARE~\cite{zhang2025FLARE} &  & 0.791 & 0.358 & 0.142 & 0.797 & 0.357 & 0.579 \\
        VGGT~\cite{wang2025vggt} &  & 0.230 & 0.678 & 0.052 & 0.969 & 0.052 & 0.968 \\
        $\pi^3$~\cite{wang2025pi} &  & \underline{0.210} & \textbf{0.726} & \underline{0.043} & \underline{0.975} & \underline{0.037} & \textbf{0.985} \\
        \textbf{Our} &  & \textbf{0.188} & \underline{0.723} & \textbf{0.037} & \textbf{0.980} & \textbf{0.037} & \underline{0.983} \\
        \bottomrule
    \end{tabular}
    \label{tab:videodepth}
\end{table}

\subsection{Comparison to the state-of-the-art}
We evaluate our method on five tasks: 3D tracking, camera pose estimation, point map estimation, video depth estimation, and monocular depth estimation. We also provide qualitative comparisons on in-the-wild field scenes in \Cref{fig:qualitative}, where our method produces cleaner geometry with fewer overlaps and better detail. In addition, we conduct comprehensive ablation studies to analyze the impact of key design.
\paragraph{\textbf{3D Point Tracking.}}
For a fair evaluation, we report results from a tracking-specific variant of our model where we freeze the pretrained backbone and fine-tune only the tracking branch. This is because our backbone is trained with a geometry-first objective and a first-frame-visible query design, which differs from the standard training/evaluation setup of 3D point trackers. As a result, directly evaluating our full model end-to-end under the geometry-oriented setup would provide an underestimate of the tracking capability enabled by our representation. We evaluate 3D tracking on TAPVid-3D~\cite{tapvid3d}, which contains large-scale real-world videos with long-range motion, occlusions, and mixed rigid/non-rigid dynamics. We report AJ$_{\text{3D}}$, APD$_{\text{3D}}$, and OA (Table~\ref{tab:tapvid3d}), which measure 3D trajectory accuracy, 3D position accuracy, and visibility prediction accuracy, respectively. We compare against (i) modular pipelines that lift 2D tracks to 3D using external depth (e.g., TAPIR~\cite{doersch2023tapir} or CoTracker3~\cite{cotracker3} with COLMAP~\cite{schonberger2016structure}, UniDepth~\cite{piccinelli2024unidepth}, or MegaSAM~\cite{li2025megasam}), and (ii) native 3D trackers that couple tracking with geometry and pose, such as TAPIP3D~\cite{tapip3d} and SpatialTracker/SpatialTracker2~\cite{SpatialTracker,SpatialTracker2}. \Cref{tab:tapvid3d} shows that our method achieves the best overall performance across subsets and on average.

\paragraph{\textbf{Camera Pose Estimation.}}
We are following recent feed-forward geometry models~\cite{wang2023posediffusion, wang2025vggt, dust3r_cvpr24, wang2025continuous, zhang2024monst3r, zhao2022particlesfm} to evaluate camera pose on RealEstate10K~\cite{zhou2018stereo} and Co3Dv2~\cite{reizenstein21co3d} using the angular-accuracy and distance error protocol adopted by recent feed-forward geometry models. For each test sequence, we sample a fixed-size subset of frames and evaluate all frame pairs. We report relative rotation accuracy (RRA), relative translation accuracy (RTA), and the AUC of the accuracy--threshold curve. We further evaluate camera trajectory accuracy on Sintel~\cite{Butler}, TUM-dynamics~\cite{6385773}, and ScanNet~\cite{dai2017scannet} with standard trajectory metrics (ATE, RPE translation, RPE rotation). \Cref{tab:relpose-vggt} presents relative pose results. On RealEstate10K, our method sets the new state-of-the-art in AUC@30 and RRA@30. On Co3Dv2, we achieve the highest rotation accuracy (RRA@30) and comparable performance on translation and AUC metrics relative to leading methods like VGGT and $\pi^3$. \Cref{tab:relpose-cut3r} reports trajectory accuracy. Our method shows strong generalization. It ranks first on TUM-dynamics across all metrics and achieves the lowest ATE on ScanNet. Notably, on Sintel, we attain the lowest RPE for both translation and rotation. While ATE on Sintel is slightly higher than $\pi^3$, our method consistently minimizes local pose errors.

\begin{table}
    \centering
    \caption{
        \textbf{Monocular Depth Estimation on Sintel~\cite{Butler}, Bonn~\cite{palazzolo2019refusion}, KITTI~\cite{geiger2013vision} and NYU-v2~\cite{silberman2012indoor}.} 
    }
    \resizebox{1.0\columnwidth}!{
    \begin{tabular}{lcccccccc}
        \toprule
        \multirow{3}{*}{\textbf{Method}} &
        \multicolumn{2}{c}{\textbf{Sintel}} &
        \multicolumn{2}{c}{\textbf{Bonn}} &
        \multicolumn{2}{c}{\textbf{KITTI}} &
        \multicolumn{2}{c}{\textbf{NYU-v2}} \\
        \cmidrule(r){2-3} \cmidrule(r){4-5} \cmidrule(r){6-7} \cmidrule(r){8-9}
        &
        Abs Rel$\downarrow$ & $\delta<1.25\uparrow$ &
        Abs Rel$\downarrow$ & $\delta<1.25\uparrow$ &
        Abs Rel$\downarrow$ & $\delta<1.25\uparrow$ &
        Abs Rel$\downarrow$ & $\delta<1.25\uparrow$ \\
        \midrule%
        DUSt3R~\cite{dust3r_cvpr24} & 0.488 & 0.532 & 0.139 & 0.832 & 0.109 & 0.873 & 0.081 & 0.909 \\
        MASt3R~\cite{mast3r_eccv24} & 0.413 & 0.569 & 0.123 & 0.833 & 0.077 & 0.948 & 0.110 & 0.865 \\
        MonST3R~\cite{zhang2024monst3r} & 0.402 & 0.525 & 0.069 & 0.954 & 0.098 & 0.895 & 0.094 & 0.887 \\
        Fast3R~\cite{Yang_2025_Fast3R} & 0.544 & 0.509 & 0.169 & 0.796 & 0.120 & 0.861 & 0.093 & 0.898 \\
        CUT3R~\cite{wang2025continuous} & 0.418 & 0.520 & 0.058 & 0.967 & 0.097 & 0.914 & 0.081 & 0.914 \\
        FLARE~\cite{zhang2025FLARE} & 0.606 & 0.402 & 0.130 & 0.836 & 0.312 & 0.513 & 0.089 & 0.898 \\
        VGGT~\cite{wang2025vggt} & 0.335 & 0.599 & 0.053 & 0.970 & 0.082 & 0.947 & 0.056 & 0.951 \\
        MoGe1~\cite{wang2025moge} & \textbf{0.273} & \textbf{0.695} & 0.050 & 0.976 & 0.054 & 0.977 & 0.055 & 0.952 \\
        MoGe2~\cite{wang2025moge2} & 0.277 & 0.687 & 0.063 & 0.973 & \textbf{0.049} & \textbf{0.979} & 0.060 & 0.940 \\
        $\pi^3$~\cite{wang2025pi} & \underline{0.277} & 0.614 & \underline{0.044} & \underline{0.976} & 0.060 & \underline{0.971} & \underline{0.054} & \underline{0.956} \\
        \textbf{Our} & 0.297 & \underline{0.617} & \textbf{0.037} & \textbf{0.979} & \underline{0.058} & 0.967 & \textbf{0.051} & \textbf{0.958} \\
        \bottomrule
    \end{tabular}
    }
    \label{tab:monodepth}
\end{table}

\begin{table}
\centering
\caption{
\textbf{Ablation Study for Joint Training on ETH3D~\cite{schoeps2017cvpr}.} 3D-Trk denote the 3D tracking branch.
}%
\footnotesize
\begin{tabular}{cccccccc}
\toprule
w. 3D-Trk &
w. $\mathcal{L}_\text{cons}$ &
w. $\mathcal{L}_\text{cam}$  &
w. $\mathcal{L}_\text{self-sup.}$ &
Acc.$\downarrow$  &
Comp.$\downarrow$ &
N.C$\uparrow$ \\
\midrule
\ding{55} & \ding{55} & \ding{55} & \ding{55} & 0.177 & 0.201 & 0.888 \\
\ding{51} & \ding{55} & \ding{55} & \ding{55} & 0.175 & 0.199 & 0.887 \\
\ding{51} & \ding{55} & \ding{51} & \ding{51} & 0.172 & 0.198 & \underline{0.890}   \\ 
\ding{51} & \ding{51} & \ding{55} & \ding{51} & 0.173 & 0.200 & 0.887   \\
\ding{51} & \ding{51} & \ding{51} & \ding{55} & \underline{0.170} & \underline{0.196} & 0.889\\
\ding{51} & \ding{51} & \ding{51} & \ding{51} & \textbf{0.168} & \textbf{0.192} & \textbf{0.891} \\
\bottomrule
\end{tabular}
\label{tab:ab1}
\end{table}

\paragraph{\textbf{Point Map Estimation.}}
We are following~\cite{wang2025continuous} to evaluate point map estimation on DTU~\cite{6909453} and ETH3D~\cite{schoeps2017cvpr}, and on long sequences from 7-Scenes~\cite{shotton2013scene} and NRGBD~\cite{azinovic2022neural} under both sparse-view and dense-view sampling settings. Following the established evaluation in recent feed-forward reconstruction work, predicted point maps are first aligned to ground truth with a similarity transform (Umeyama), then optionally refined with ICP; we report Accuracy, Completion, and Normal Consistency. 

As reported in \Cref{tab:mv_recon_vggt}, our method demonstrates strong generalization across different scene types. Our method significantly outperforms all baselines on the complex, scene-centric ETH3D dataset, achieving the best performance across Accuracy, Completion, and Normal Consistency metrics. Furthermore, \Cref{tab:mv_recon_cut3r} highlights the robustness of our approach on long video sequences. On 7-Scenes, our method ranks first in both sparse-view and dense-view settings, surpassing the recent state-of-the-art $\pi^3$. On NRGBD, we achieve results comparable to the top-performing method, confirming that our point map estimation remains stable and accurate even over extended trajectories.

\paragraph{\textbf{Video Depth Estimation.}}
We are following~\cite{wang2025continuous, zhang2024monst3r} to evaluate video depth on Sintel~\cite{Butler}, Bonn~\cite{palazzolo2019refusion}, and KITTI~\cite{geiger2013vision}. We follow the common protocol of evaluating depth under (i) a single scale alignment per sequence and (ii) a shared scale-and-shift alignment per sequence, and report Abs Rel and threshold accuracy. Quantitative results are presented in \Cref{tab:videodepth}. Our method sets a new state-of-the-art for video depth estimation, particularly excelling in dynamic environments. Under the scale-only alignment protocol, we achieve the lowest Abs Rel errors on Sintel (0.220) and Bonn (0.036), outperforming $\pi^3$ by a clear margin. This advantage holds under the scale-and-shift protocol, where our method ranks first on Sintel and ties for the best performance on KITTI. 

\begin{table}
    \centering
    \caption{\textbf{Ablation Study for semi-supervised training on Sekai~\cite{li2025sekai}.}}
    \begin{tabular}{lccc}
        \toprule
        {Setting} & ATE$\downarrow$ & RPE trans$\downarrow$ & RPE rot$\downarrow$ \\
        \midrule

        w. $\mathcal{L}_\text{self-sup.}$
        & 0.0101 & 0.0051 & 0.2515 \\

        wo. $\mathcal{L}_\text{self-sup.}$
        & 0.0154 & 0.0103 & 0.4413 \\

        \bottomrule
    \end{tabular}
    \label{tab:ab2}
\end{table}

\paragraph{\textbf{Monocular Depth Estimation.}}
We are following~\cite{wang2025continuous, zhang2024monst3r} to evaluate monocular depth on Sintel~\cite{Butler}, Bonn~\cite{palazzolo2019refusion}, KITTI~\cite{geiger2013vision}, and NYU-v2~\cite{silberman2012indoor}. Following prior work, each predicted depth map is aligned independently to ground truth (per-image scale), and we report Abs Rel and threshold accuracy. \Cref{tab:monodepth} shows the zero-shot evaluation on monocular depth benchmarks. We achieve the best performance on Bonn and NYU-v2, surpassing both baseline methods (e.g., $\pi^3$) and specialized monocular estimators (e.g., MoGe). On Sintel and KITTI, our method remains highly competitive, ranking second-best overall. 

\begin{figure}
    \centering
    \includegraphics[width=0.6\linewidth]{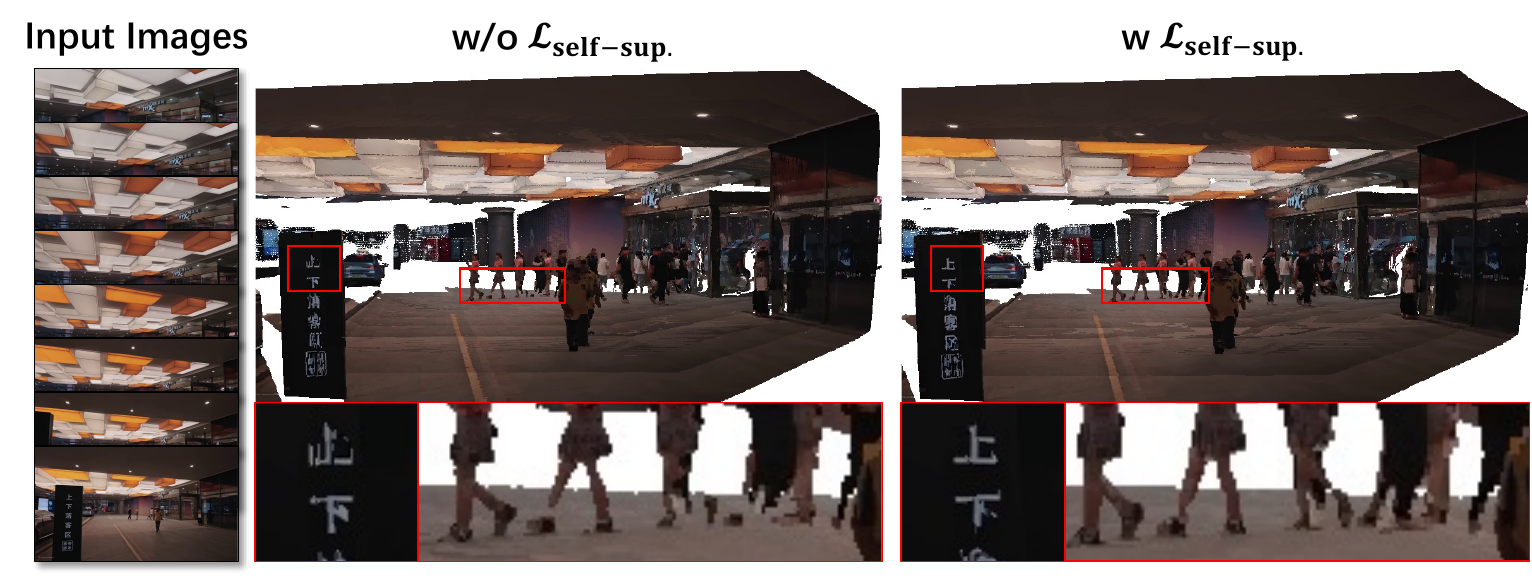}
    \caption{\textbf{The influence of semi-supervised training.}}
    \label{fig:ab1}
\end{figure}

\begin{table}
    \centering
    \caption{\textbf{Ablation study for tracking branch on DynamicReplica~\cite{DynamicReplica}.}}
    \begin{tabular}{l|ccc}
        \toprule
        \textbf{Setting} & AJ$_{\text{3D}}\uparrow$ & APD$_{\text{3D}}\uparrow$ & OA$\uparrow$ \\
        \midrule
        Base (Only 3D track branch) & 49.8 & 65.7 & 83.4 \\
        + $\mathcal{L}_\text{cons}$ & 57.4 & 70.1 & 84.1 \\
        + $\mathcal{L}_\text{cons}$ + $\mathcal{L}_\text{cam}$  & \underline{59.1} & \underline{72.3} & \underline{85.0} \\
        + $\mathcal{L}_\text{cons}$ + $\mathcal{L}_\text{cam}$ + $\mathcal{L}_\text{self-sup.}$ & \textbf{59.5} & \textbf{72.5} & \textbf{85.1} \\
        \bottomrule
    \end{tabular}
    \label{tab:ab3}
\end{table}

\subsection{Ablation Study}
The ablation study investigates the impact of joint training and self-supervision training. \Cref{tab:ab1} shows the effect of adding our 3D tracking branch and the associated losses on ETH3D. Enabling the 3D tracking branch alone already improves point-map estimation, suggesting that 3D tracking acts as an auxiliary task that strengthens the shared representation for geometry prediction. This is in line with the observation in VGGT~\cite{wang2025vggt} that jointly predicting multiple 3D quantities benefits point-map quality. When we further introduce the camera consistency term $\mathcal{L}_\text{cam}$ together with the pointmap--tracking bidirectional consistency term $\mathcal{L}_\text{cons}$, the improvement becomes clearly larger than using either term in isolation, indicating that accurate cross-view pose alignment and dense-sparse coupling provide complementary constraints for multi-frame fusion. Finally, adding the self-supervised term $\mathcal{L}_\text{self-sup.}$ only brings a modest additional gain on ETH3D. We attribute this mainly to the data distribution gap, since the self-supervised stage is trained on in-the-wild Sekai videos~\cite{li2025sekai} while evaluation is performed on ETH3D. Therefore, we further investigate the self-supervision effect in an additional ablation on Sekai itself. As shown in \Cref{tab:ab2}, incorporating $\mathcal{L}_\text{self-sup.}$ consistently improves both global trajectory accuracy and relative pose errors, and \Cref{fig:ab1} improves detail for the pointmap. It confirms that our self-supervision primarily enhances performance on in-the-wild data. 

In addition, we analyze how the tracking-related losses affect 3D tracking quality on Dynamic Replica. \Cref{tab:ab3} starts from the base model with only the 3D tracking branch, and adding $\mathcal{L}_\text{cons}$ yields a clear improvement across all three metrics, demonstrating the benefit of enforcing consistency between point maps and sparse tracks. Adding $\mathcal{L}_\text{cam}$ further improves performance, indicating that better camera alignment also helps tracking in dynamic scenes. Finally, including $\mathcal{L}_\text{self-sup.}$ provides a small but consistent additional gain and achieves the best overall results.

\section{Conclusion}
We present a feed-forward multi-frame reconstruction pipeline that improves geometry and camera pose by making 3D correspondence an explicit prediction. The model predicts camera-space 3D trajectories and couples them with dense local point maps and relative camera poses through bidirectional track–pointmap consistency and a pose objective driven by static track anchors, while suppressing gradients from dynamic regions. This coupling turns correspondence errors into direct training signals for both geometry and pose.  We also translate the same coupling constraints into a semi-supervised form that only needs 2D tracks. We demonstrate that this coupled design is effective in practice and consistently improves reconstruction quality and camera motion estimation over strong feed-forward baselines. 

%Bibliography
\bibliographystyle{unsrt}  
\bibliography{references}

\begin{thebibliography}{10}

\bibitem{hartley2003multiple}
Richard Hartley and Andrew Zisserman.
\newblock {\em Multiple view geometry in computer vision}.
\newblock Cambridge university press, 2003.

\bibitem{triggs1999bundle}
Bill Triggs, Philip~F McLauchlan, Richard~I Hartley, and Andrew~W Fitzgibbon.
\newblock Bundle adjustment—a modern synthesis.
\newblock In {\em International workshop on vision algorithms}, pages 298--372. Springer, 1999.

\bibitem{lowe2004distinctive}
David~G Lowe.
\newblock Distinctive image features from scale-invariant keypoints.
\newblock {\em International journal of computer vision}, 60(2):91--110, 2004.

\bibitem{schonberger2016structure}
Johannes~L Schonberger and Jan-Michael Frahm.
\newblock Structure-from-motion revisited.
\newblock In {\em Proceedings of the IEEE conference on computer vision and pattern recognition}, pages 4104--4113, 2016.

\bibitem{agarwal2011building}
Sameer Agarwal, Yasutaka Furukawa, Noah Snavely, Ian Simon, Brian Curless, Steven~M Seitz, and Richard Szeliski.
\newblock Building rome in a day.
\newblock {\em Communications of the ACM}, 54(10):105--112, 2011.

\bibitem{teed2021droid}
Zachary Teed and Jia Deng.
\newblock Droid-slam: Deep visual slam for monocular, stereo, and rgb-d cameras.
\newblock {\em Advances in neural information processing systems}, 34:16558--16569, 2021.

\bibitem{zhou2017unsupervised}
Tinghui Zhou, Matthew Brown, Noah Snavely, and David~G Lowe.
\newblock Unsupervised learning of depth and ego-motion from video.
\newblock In {\em Proceedings of the IEEE conference on computer vision and pattern recognition}, pages 1851--1858, 2017.

\bibitem{dust3r_cvpr24}
Shuzhe Wang, Vincent Leroy, Yohann Cabon, Boris Chidlovskii, and Jerome Revaud.
\newblock Dust3r: Geometric 3d vision made easy.
\newblock In {\em CVPR}, 2024.

\bibitem{murai2025mast3r}
Riku Murai, Eric Dexheimer, and Andrew~J Davison.
\newblock Mast3r-slam: Real-time dense slam with 3d reconstruction priors.
\newblock In {\em Proceedings of the Computer Vision and Pattern Recognition Conference}, pages 16695--16705, 2025.

\bibitem{fang2025dens3r}
Xianze Fang, Jingnan Gao, Zhe Wang, Zhuo Chen, Xingyu Ren, Jiangjing Lyu, Qiaomu Ren, Zhonglei Yang, Xiaokang Yang, Yichao Yan, et~al.
\newblock Dens3r: A foundation model for 3d geometry prediction.
\newblock {\em arXiv preprint arXiv:2507.16290}, 2025.

\bibitem{wang2025vggt}
Jianyuan Wang, Minghao Chen, Nikita Karaev, Andrea Vedaldi, Christian Rupprecht, and David Novotny.
\newblock Vggt: Visual geometry grounded transformer.
\newblock In {\em Proceedings of the IEEE/CVF Conference on Computer Vision and Pattern Recognition}, 2025.

\bibitem{wang2025pi}
Yifan Wang, Jianjun Zhou, Haoyi Zhu, Wenzheng Chang, Yang Zhou, Zizun Li, Junyi Chen, Jiangmiao Pang, Chunhua Shen, and Tong He.
\newblock $\pi^3$: Permutation-equivariant visual geometry learning.
\newblock {\em arXiv preprint arXiv:2507.13347}, 2025.

\bibitem{st4rtrack2025}
Haiwen Feng*, Junyi Zhang*, Qianqian Wang, Yufei Ye, Pengcheng Yu, Michael~J. Black, Trevor Darrell, and Angjoo Kanazawa.
\newblock St4rtrack: Simultaneous 4d reconstruction and tracking in the world.
\newblock In {\em Proceedings of the IEEE/CVF International Conference on Computer Vision}, 2025.

\bibitem{mast3r_eccv24}
Vincent Leroy, Yohann Cabon, and Jerome Revaud.
\newblock Grounding image matching in 3d with mast3r, 2024.

\bibitem{cotracker2}
Nikita Karaev, Ignacio Rocco, Benjamin Graham, Natalia Neverova, Andrea Vedaldi, and Christian Rupprecht.
\newblock {CoTracker}: It is better to track together.
\newblock 2023.

\bibitem{cotracker3}
Nikita Karaev, Iurii Makarov, Jianyuan Wang, Natalia Neverova, Andrea Vedaldi, and Christian Rupprecht.
\newblock {CoTracker3}: Simpler and better point tracking by pseudo-labelling real videos.
\newblock 2024.

\bibitem{harley2022particle}
Adam~W Harley, Zhaoyuan Fang, and Katerina Fragkiadaki.
\newblock Particle video revisited: Tracking through occlusions using point trajectories.
\newblock In {\em ECCV}, 2022.

\bibitem{doersch2023tapir}
Carl Doersch, Yi~Yang, Mel Vecerik, Dilara Gokay, Ankush Gupta, Yusuf Aytar, Joao Carreira, and Andrew Zisserman.
\newblock {TAPIR}: Tracking any point with per-frame initialization and temporal refinement.
\newblock In {\em Proceedings of the IEEE/CVF International Conference on Computer Vision}, pages 10061--10072, 2023.

\bibitem{tapvid3d}
Skanda Koppula, Ignacio Rocco, Yi~Yang, Joe Heyward, João Carreira, Andrew Zisserman, Gabriel Brostow, and Carl Doersch.
\newblock Tapvid-3d: A benchmark for tracking any point in 3d, 2024.

\bibitem{wang2023omnimotion}
Qianqian Wang, Yen-Yu Chang, Ruojin Cai, Zhengqi Li, Bharath Hariharan, Aleksander Holynski, and Noah Snavely.
\newblock Tracking everything everywhere all at once.
\newblock {\em ICCV}, 2023.

\bibitem{ngo2024delta}
Tuan~Duc Ngo, Peiye Zhuang, Chuang Gan, Evangelos Kalogerakis, Sergey Tulyakov, Hsin-Ying Lee, and Chaoyang Wang.
\newblock Delta: Dense efficient long-range 3d tracking for any video.
\newblock {\em arXiv preprint arXiv:2410.24211}, 2024.

\bibitem{rajic2025mvtracker}
Frano Raji{\v{c}}, Haofei Xu, Marko Mihajlovic, Siyuan Li, Irem Demir, Emircan G{\"u}ndo{\u{g}}du, Lei Ke, Sergey Prokudin, Marc Pollefeys, and Siyu Tang.
\newblock Multi-view 3d point tracking.
\newblock In {\em Proceedings of the IEEE/CVF International Conference on Computer Vision (ICCV)}, 2025.

\bibitem{SpatialTracker}
Yuxi Xiao, Qianqian Wang, Shangzhan Zhang, Nan Xue, Sida Peng, Yujun Shen, and Xiaowei Zhou.
\newblock Spatialtracker: Tracking any 2d pixels in 3d space.
\newblock In {\em Proceedings of the IEEE/CVF Conference on Computer Vision and Pattern Recognition (CVPR)}, 2024.

\bibitem{tapip3d}
Bowei Zhang, Lei Ke, Adam~W Harley, and Katerina Fragkiadaki.
\newblock Tapip3d: Tracking any point in persistent 3d geometry.
\newblock {\em arXiv preprint arXiv:2504.14717}, 2025.

\bibitem{SpatialTracker2}
Yuxi Xiao, Jianyuan Wang, Nan Xue, Nikita Karaev, Iurii Makarov, Bingyi Kang, Xin Zhu, Hujun Bao, Yujun Shen, and Xiaowei Zhou.
\newblock Spatialtrackerv2: 3d point tracking made easy.
\newblock In {\em ICCV}, 2025.

\bibitem{dai2017scannet}
Angela Dai, Angel~X. Chang, Manolis Savva, Maciej Halber, Thomas Funkhouser, and Matthias Nie{\ss}ner.
\newblock Scannet: Richly-annotated 3d reconstructions of indoor scenes.
\newblock In {\em Proc. Computer Vision and Pattern Recognition (CVPR), IEEE}, 2017.

\bibitem{reizenstein21co3d}
Jeremy Reizenstein, Roman Shapovalov, Philipp Henzler, Luca Sbordone, Patrick Labatut, and David Novotny.
\newblock Common objects in 3d: Large-scale learning and evaluation of real-life 3d category reconstruction.
\newblock In {\em International Conference on Computer Vision}, 2021.

\bibitem{piccinelli2024unidepth}
Luigi Piccinelli, Yung-Hsu Yang, Christos Sakaridis, Mattia Segu, Siyuan Li, Luc Van~Gool, and Fisher Yu.
\newblock Unidepth: Universal monocular metric depth estimation.
\newblock In {\em Proceedings of the IEEE/CVF Conference on Computer Vision and Pattern Recognition}, pages 10106--10116, 2024.

\bibitem{li2025megasam}
Zhengqi Li, Richard Tucker, Forrester Cole, Qianqian Wang, Linyi Jin, Vickie Ye, Angjoo Kanazawa, Aleksander Holynski, and Noah Snavely.
\newblock Megasam: Accurate, fast and robust structure and motion from casual dynamic videos.
\newblock In {\em Proceedings of the Computer Vision and Pattern Recognition Conference}, pages 10486--10496, 2025.

\bibitem{delta2}
Tuan~Duc Ngo, Ashkan Mirzaei, Guocheng Qian, Hanwen Liang, Chuang Gan, Evangelos Kalogerakis, Peter Wonka, and Chaoyang Wang.
\newblock Deltav2: Accelerating dense 3d tracking, 2025.

\bibitem{zhou2018stereo}
Tinghui Zhou, Richard Tucker, John Flynn, Graham Fyffe, and Noah Snavely.
\newblock Stereo magnification: Learning view synthesis using multiplane images.
\newblock In {\em SIGGRAPH}, 2018.

\bibitem{Yang_2025_Fast3R}
Jianing Yang, Alexander Sax, Kevin~J. Liang, Mikael Henaff, Hao Tang, Ang Cao, Joyce Chai, Franziska Meier, and Matt Feiszli.
\newblock Fast3r: Towards 3d reconstruction of 1000+ images in one forward pass.
\newblock In {\em Proceedings of the IEEE/CVF Conference on Computer Vision and Pattern Recognition (CVPR)}, June 2025.

\bibitem{wang2025continuous}
Qianqian Wang, Yifei Zhang, Aleksander Holynski, Alexei~A Efros, and Angjoo Kanazawa.
\newblock Continuous 3d perception model with persistent state.
\newblock {\em arXiv preprint arXiv:2501.12387}, 2025.

\bibitem{zhang2025FLARE}
Shangzhan Zhang, Jianyuan Wang, Yinghao Xu, Nan Xue, Christian Rupprecht, Xiaowei Zhou, Yujun Shen, and Gordon Wetzstein.
\newblock Flare: Feed-forward geometry, appearance and camera estimation from uncalibrated sparse views, 2025.

\bibitem{arnold2022mapfree}
Eduardo Arnold, Jamie Wynn, Sara Vicente, Guillermo Garcia-Hernando, {\'{A}}ron Monszpart, Victor~Adrian Prisacariu, Daniyar Turmukhambetov, and Eric Brachmann.
\newblock Map-free visual relocalization: Metric pose relative to a single image.
\newblock In {\em ECCV}, 2022.

\bibitem{tartanair2020iros}
Wenshan Wang, Delong Zhu, Xiangwei Wang, Yaoyu Hu, Yuheng Qiu, Chen Wang, Yafei Hu, Ashish Kapoor, and Sebastian Scherer.
\newblock Tartanair: A dataset to push the limits of visual slam.
\newblock 2020.

\bibitem{avetisyan2024scenescript}
Armen Avetisyan, Christopher Xie, Henry Howard-Jenkins, Tsun-Yi Yang, Samir Aroudj, Suvam Patra, Fuyang Zhang, Duncan Frost, Luke Holland, Campbell Orme, Jakob Engel, Edward Miller, Richard Newcombe, and Vasileios Balntas.
\newblock Scenescript: Reconstructing scenes with an autoregressive structured language model.
\newblock In {\em European Conference on Computer Vision (ECCV)}, 2024.

\bibitem{cabon2020vkitti2}
Yohann Cabon, Naila Murray, and Martin Humenberger.
\newblock Virtual kitti 2, 2020.

\bibitem{DeepMVS}
Po-Han Huang, Kevin Matzen, Johannes Kopf, Narendra Ahuja, and Jia-Bin Huang.
\newblock Deepmvs: Learning multi-view stereopsis.
\newblock In {\em IEEE Conference on Computer Vision and Pattern Recognition (CVPR)}, 2018.

\bibitem{dehghan2021arkitscenes}
Gilad Baruch, Zhuoyuan Chen, Afshin Dehghan, Tal Dimry, Yuri Feigin, Peter Fu, Thomas Gebauer, Brandon Joffe, Daniel Kurz, Arik Schwartz, and Elad Shulman.
\newblock {ARK}itscenes - a diverse real-world dataset for 3d indoor scene understanding using mobile {RGB}-d data.
\newblock In {\em Thirty-fifth Conference on Neural Information Processing Systems Datasets and Benchmarks Track (Round 1)}, 2021.

\bibitem{yao2020blendedmvs}
Yao Yao, Zixin Luo, Shiwei Li, Jingyang Zhang, Yufan Ren, Lei Zhou, Tian Fang, and Long Quan.
\newblock Blendedmvs: A large-scale dataset for generalized multi-view stereo networks.
\newblock {\em Computer Vision and Pattern Recognition (CVPR)}, 2020.

\bibitem{ling2024dl3dv}
Lu~Ling, Yichen Sheng, Zhi Tu, Wentian Zhao, Cheng Xin, Kun Wan, Lantao Yu, Qianyu Guo, Zixun Yu, Yawen Lu, et~al.
\newblock Dl3dv-10k: A large-scale scene dataset for deep learning-based 3d vision.
\newblock In {\em Proceedings of the IEEE/CVF Conference on Computer Vision and Pattern Recognition}, pages 22160--22169, 2024.

\bibitem{yeshwanthliu2023scannetpp}
Chandan Yeshwanth, Yueh-Cheng Liu, Matthias Nie{\ss}ner, and Angela Dai.
\newblock Scannet++: A high-fidelity dataset of 3d indoor scenes.
\newblock In {\em Proceedings of the International Conference on Computer Vision ({ICCV})}, 2023.

\bibitem{MegaDepthLi18}
Zhengqi Li and Noah Snavely.
\newblock Megadepth: Learning single-view depth prediction from internet photos.
\newblock In {\em Computer Vision and Pattern Recognition (CVPR)}, 2018.

\bibitem{Waymo}
Pei Sun, Henrik Kretzschmar, Xerxes Dotiwalla, Aurelien Chouard, Vijaysai Patnaik, Paul Tsui, James Guo, Yin Zhou, Yuning Chai, Benjamin Caine, Vijay Vasudevan, Wei Han, Jiquan Ngiam, Hang Zhao, Aleksei Timofeev, Scott Ettinger, Maxim Krivokon, Amy Gao, Aditya Joshi, Sheng Zhao, Shuyang Cheng, Yu~Zhang, Jonathon Shlens, Zhifeng Chen, and Dragomir Anguelov.
\newblock Scalability in perception for autonomous driving: Waymo open dataset, 2020.

\bibitem{xia2024rgbd}
Hongchi Xia, Yang Fu, Sifei Liu, and Xiaolong Wang.
\newblock Rgbd objects in the wild: Scaling real-world 3d object learning from rgb-d videos, 2024.

\bibitem{GTASfm}
Kaixuan Wang and Shaojie Shen.
\newblock Flow-motion and depth network for monocular stereo and beyond, 2019.

\bibitem{Hypersim}
Mike Roberts, Jason Ramapuram, Anurag Ranjan, Atulit Kumar, Miguel~Angel Bautista, Nathan Paczan, Russ Webb, and Joshua~M. Susskind.
\newblock {Hypersim}: {A} photorealistic synthetic dataset for holistic indoor scene understanding.
\newblock In {\em International Conference on Computer Vision (ICCV) 2021}, 2021.

\bibitem{zhou2025omniworld}
Yang Zhou, Yifan Wang, Jianjun Zhou, Wenzheng Chang, Haoyu Guo, Zizun Li, Kaijing Ma, Xinyue Li, Yating Wang, Haoyi Zhu, Mingyu Liu, Dingning Liu, Jiange Yang, Zhoujie Fu, Junyi Chen, Chunhua Shen, Jiangmiao Pang, Kaipeng Zhang, and Tong He.
\newblock Omniworld: A multi-domain and multi-modal dataset for 4d world modeling.
\newblock {\em arXiv preprint arXiv:2509.12201}, 2025.

\bibitem{aleotti2021neural}
Filippo Aleotti, Fabio Tosi, Pierluigi Zama~Ramirez, Matteo Poggi, Samuele Salti, Luigi Di~Stefano, and Stefano Mattoccia.
\newblock Neural disparity refinement for arbitrary resolution stereo.
\newblock In {\em International Conference on 3D Vision}, 2021.
\newblock 3DV.

\bibitem{li2023matrixcity}
Yixuan Li, Lihan Jiang, Linning Xu, Yuanbo Xiangli, Zhenzhi Wang, Dahua Lin, and Bo~Dai.
\newblock Matrixcity: A large-scale city dataset for city-scale neural rendering and beyond.
\newblock In {\em Proceedings of the IEEE/CVF International Conference on Computer Vision}, pages 3205--3215, 2023.

\bibitem{Mehl2023_Spring}
Lukas Mehl, Jenny Schmalfuss, Azin Jahedi, Yaroslava Nalivayko, and Andr\'es Bruhn.
\newblock Spring: A high-resolution high-detail dataset and benchmark for scene flow, optical flow and stereo.
\newblock In {\em Proc. IEEE/CVF Conference on Computer Vision and Pattern Recognition (CVPR)}, 2023.

\bibitem{greff2021kubric}
Klaus Greff, Francois Belletti, Lucas Beyer, Carl Doersch, Yilun Du, Daniel Duckworth, David~J Fleet, Dan Gnanapragasam, Florian Golemo, Charles Herrmann, Thomas Kipf, Abhijit Kundu, Dmitry Lagun, Issam Laradji, Hsueh-Ti~(Derek) Liu, Henning Meyer, Yishu Miao, Derek Nowrouzezahrai, Cengiz Oztireli, Etienne Pot, Noha Radwan, Daniel Rebain, Sara Sabour, Mehdi S.~M. Sajjadi, Matan Sela, Vincent Sitzmann, Austin Stone, Deqing Sun, Suhani Vora, Ziyu Wang, Tianhao Wu, Kwang~Moo Yi, Fangcheng Zhong, and Andrea Tagliasacchi.
\newblock Kubric: a scalable dataset generator.
\newblock 2022.

\bibitem{zheng2023point}
Yang Zheng, Adam~W. Harley, Bokui Shen, Gordon Wetzstein, and Leonidas~J. Guibas.
\newblock Pointodyssey: A large-scale synthetic dataset for long-term point tracking.
\newblock In {\em ICCV}, 2023.

\bibitem{DynamicReplica}
Nikita Karaev, Ignacio Rocco, Benjamin Graham, Natalia Neverova, Andrea Vedaldi, and Christian Rupprecht.
\newblock Dynamicstereo: Consistent dynamic depth from stereo videos.
\newblock {\em CVPR}, 2023.

\bibitem{Butler}
D.~J. Butler, J.~Wulff, G.~B. Stanley, and M.~J. Black.
\newblock A naturalistic open source movie for optical flow evaluation.
\newblock In {A. Fitzgibbon et al. (Eds.)}, editor, {\em European Conf. on Computer Vision (ECCV)}, Part IV, LNCS 7577, pages 611--625. Springer-Verlag, October 2012.

\bibitem{6385773}
Jürgen Sturm, Nikolas Engelhard, Felix Endres, Wolfram Burgard, and Daniel Cremers.
\newblock A benchmark for the evaluation of rgb-d slam systems.
\newblock In {\em 2012 IEEE/RSJ International Conference on Intelligent Robots and Systems}, pages 573--580, 2012.

\bibitem{aether}
Aether Team, Haoyi Zhu, Yifan Wang, Jianjun Zhou, Wenzheng Chang, Yang Zhou, Zizun Li, Junyi Chen, Chunhua Shen, Jiangmiao Pang, and Tong He.
\newblock Aether: Geometric-aware unified world modeling.
\newblock {\em arXiv preprint arXiv:2503.18945}, 2025.

\bibitem{6909453}
Rasmus Jensen, Anders Dahl, George Vogiatzis, Engil Tola, and Henrik Aanæs.
\newblock Large scale multi-view stereopsis evaluation.
\newblock In {\em 2014 IEEE Conference on Computer Vision and Pattern Recognition}, pages 406--413, 2014.

\bibitem{schoeps2017cvpr}
Thomas Sch\"ops, Johannes~L. Sch\"onberger, Silvano Galliani, Torsten Sattler, Konrad Schindler, Marc Pollefeys, and Andreas Geiger.
\newblock A multi-view stereo benchmark with high-resolution images and multi-camera videos.
\newblock In {\em Conference on Computer Vision and Pattern Recognition (CVPR)}, 2017.

\bibitem{shotton2013scene}
Jamie Shotton, Ben Glocker, Christopher Zach, Shahram Izadi, Antonio Criminisi, and Andrew Fitzgibbon.
\newblock Scene coordinate regression forests for camera relocalization in {RGB-D} images.
\newblock In {\em CVPR}, 2013.

\bibitem{azinovic2022neural}
Dejan Azinovi\'c, Ricardo Martin-Brualla, Dan~B Goldman, Matthias Nie{\ss}ner, and Justus Thies.
\newblock Neural rgb-d surface reconstruction.
\newblock In {\em Proceedings of the IEEE/CVF Conference on Computer Vision and Pattern Recognition (CVPR)}, pages 6290--6301, June 2022.

\bibitem{palazzolo2019refusion}
Emanuele Palazzolo, Jens Behley, Philipp Lottes, Philippe Giguère, and Cyrill Stachniss.
\newblock Refusion: 3d reconstruction in dynamic environments for rgb-d cameras exploiting residuals, 2019.

\bibitem{geiger2013vision}
Andreas Geiger, Philip Lenz, Christoph Stiller, and Raquel Urtasun.
\newblock Vision meets robotics: The kitti dataset.
\newblock {\em The International Journal of Robotics Research}, 32(11):1231--1237, 2013.

\bibitem{zhang2024monst3r}
Junyi Zhang, Charles Herrmann, Junhwa Hur, Varun Jampani, Trevor Darrell, Forrester Cole, Deqing Sun, and Ming-Hsuan Yang.
\newblock Monst3r: A simple approach for estimating geometry in the presence of motion.
\newblock {\em arXiv preprint arxiv:2410.03825}, 2024.

\bibitem{tang2024mv}
Zhenggang Tang, Yuchen Fan, Dilin Wang, Hongyu Xu, Rakesh Ranjan, Alexander Schwing, and Zhicheng Yan.
\newblock Mv-dust3r+: Single-stage scene reconstruction from sparse views in 2 seconds.
\newblock {\em arXiv preprint arXiv:2412.06974}, 2024.

\bibitem{wang2023posediffusion}
Jianyuan Wang, Christian Rupprecht, and David Novotny.
\newblock Posediffusion: Solving pose estimation via diffusion-aided bundle adjustment.
\newblock In {\em Proceedings of the IEEE/CVF International Conference on Computer Vision}, pages 9773--9783, 2023.

\bibitem{zhao2022particlesfm}
Wang Zhao, Shaohui Liu, Hengkai Guo, Wenping Wang, and Yong-Jin Liu.
\newblock Particlesfm: Exploiting dense point trajectories for localizing moving cameras in the wild.
\newblock In {\em European Conference on Computer Vision}, pages 523--542. Springer, 2022.

\bibitem{silberman2012indoor}
Pushmeet~Kohli Nathan~Silberman, Derek~Hoiem and Rob Fergus.
\newblock Indoor segmentation and support inference from rgbd images.
\newblock In {\em ECCV}, 2012.

\bibitem{wang2025moge}
Ruicheng Wang, Sicheng Xu, Cassie Dai, Jianfeng Xiang, Yu~Deng, Xin Tong, and Jiaolong Yang.
\newblock Moge: Unlocking accurate monocular geometry estimation for open-domain images with optimal training supervision.
\newblock In {\em Proceedings of the Computer Vision and Pattern Recognition Conference}, pages 5261--5271, 2025.

\bibitem{wang2025moge2}
Ruicheng Wang, Sicheng Xu, Yue Dong, Yu~Deng, Jianfeng Xiang, Zelong Lv, Guangzhong Sun, Xin Tong, and Jiaolong Yang.
\newblock Moge-2: Accurate monocular geometry with metric scale and sharp details, 2025.

\bibitem{li2025sekai}
Zhen Li, Chuanhao Li, Xiaofeng Mao, Shaoheng Lin, Ming Li, Shitian Zhao, Zhaopan Xu, Xinyue Li, Yukang Feng, Jianwen Sun, Zizhen Li, Fanrui Zhang, Jiaxin Ai, Zhixiang Wang, Yuwei Wu, Tong He, Jiangmiao Pang, Yu~Qiao, Yunde Jia, and Kaipeng Zhang.
\newblock Sekai: A video dataset towards world exploration.
\newblock {\em arXiv preprint arXiv:2506.15675}, 2025.

\end{thebibliography}

\end{document}